\documentclass[journal]{IEEEtran}
\usepackage{cite}
\ifCLASSINFOpdf
\else
\fi
\usepackage{amsmath}

\usepackage{array}
\usepackage{url}
\usepackage{graphics}
\usepackage{graphicx}
\usepackage{epstopdf}
\usepackage{amssymb}
\usepackage{algorithm,algorithmic}
\usepackage{hyperref}
\usepackage{url}
\usepackage{multirow}
\usepackage{array}
\newcolumntype{M}[1]{>{\centering\arraybackslash}m{#1}}
\usepackage{amsmath}
\usepackage{float}
\usepackage[caption=false]{subfig}
\usepackage{color}
\usepackage{caption}
\usepackage{dsfont}
\begin{document}
\title{PoreNet:  CNN-based Pore Descriptor for High-resolution Fingerprint Recognition}

\author{Vijay~Anand,~\IEEEmembership{Student Member,~IEEE,}
        and~Vivek~Kanhangad,~\IEEEmembership{Senior Member,~IEEE}% <-this % stops a space
\thanks{V. Anand  and V. Kanhangad are with the Discipline of Electrical Engineering, Indian Institute of Technology Indore, Indore 453552, India.
 e-mail: phd1401202011@iiti.ac.in (V. Anand),  kvivek@iiti.ac.in (V. Kanhangad).}% <-this % stops a space
}

\maketitle

\begin{abstract}
With the development of high-resolution fingerprint scanners, high-resolution fingerprint-based biometric recognition has received increasing attention in recent years. This paper presents a pore feature-based approach for biometric recognition. Our approach employs a convolutional neural network (CNN) model, DeepResPore, to detect pores  in  the input fingerprint image. Thereafter, a CNN-based descriptor is computed for a patch around each detected pore. Specifically, we have designed a residual learning-based  CNN, referred  to  as PoreNet that learns distinctive feature representation from pore patches. For verification, a matching score is generated by comparing the pore descriptors, obtained from a pair of  fingerprint images, in a bi-directional manner using the Euclidean distance. %Matches between the descriptors are refined using the distance ratio threshold on the second nearest neighbour.
 The proposed approach for high-resolution fingerprint recognition achieves 2.91\% and 0.57\% equal error rates (EERs) on partial (DBI) and complete (DBII)  fingerprints of the benchmark PolyU HRF dataset. %On an in-house dataset, our approach achieves x\% and y\% EERs for partial and complete fingerprints, respectively.
Most importantly, it achieves lower FMR1000 and FMR10000 values than the current state-of-the-art approach on both the datasets. 
\end{abstract}


% \fi
\IEEEpeerreviewmaketitle

\section{Introduction}
\label{intro}
\IEEEPARstart{F}{ingerprint} is one of the most widely explored biometric traits, mainly due its distinctiveness and permanence \cite{maltoni2009handbook}.  
The features extracted from a fingerprint image are broadly classified into level-1, level-2 and level-3 features. Level-1 features, which include global ridge orientation, are commonly used for fingerprint classification. Level-2 fingerprint features include finer details such as ridge endings and ridge bifurcations, which are collectively called minutiae \cite{maltoni2009handbook}. Level-3 fingerprint features, on the other hand, include very fine details such as pores, incipient ridges, dots, and ridge contours. Level-1 and level-2 features can be observed in 500 dpi fingerprint images, while level-3 features are generally observable in  fingerprint images having a resolution greater than 800 dpi \cite{HEF_resolution}.

Commercially available automated fingerprint recognition systems (AFRS) and a majority of the methods reported in the literature employ level-1 and level-2 features. However, with the advent of high-resolution fingerprint sensors, there has been a focus shift and several methods that employ level-3 features have been developed for fingerprint recognition.  In addition to enhancing the recognition performance, level-3 features provide higher level of security, as they are difficult to forge. Further, the level-3 features have also been included in the extended  feature set for fingerprint recognition \cite{extended_feature}.  
Over the last few years, there has been growing interest in level-3 fingerprint features, especially the pores and several methods have been proposed for pore feature based automated fingerprint recognition \cite{stosz_pore,roddy1997fingerprint,kryszczuk2004extraction,kryszczuk2004study,jain2007pores, zhao2009direct, Zhao20102833, ZHAO_partial,sparse_fing, LIU_PR,Lemes,segundo,vijay_pore}. 
A pore-based AFRS typically consists of two major steps namely, pore detection in high-resolution fingerprint images and matching fingerprints using the detected pores.
Stosz and Alyea \cite{stosz_pore} in their pioneering work proposed a fingerprint recognition approach that uses both pores and minutiae.
 Their approach involves detecting pores by tracing the ridges in skeletonized fingerprint image, followed by a multi-level matching using pores and minutiae. 
Roddy and Stosz \cite{roddy1997fingerprint} provided a detailed discussion on the statistics of the pores and examined its effectiveness in improving the performance of the existing AFRS.
Krysczuk \textit{et al.} \cite{kryszczuk2004extraction,kryszczuk2004study} demonstrated the efficacy of pore features for fragmentary fingerprint recognition. In their approach, closed pores are detected by applying a set of thresholds to the binarized fingerprint image and open pores are detected by skeletonizing the valleys and finding the spurs having a sufficient number of white pixels in the neighbourhood. 
 Their experimental results demonstrated that pore features are vital in recognizing partial fingerprint.
The early studies \cite{stosz_pore,roddy1997fingerprint,kryszczuk2004extraction,kryszczuk2004study} employed skeletonization-based approaches to detect pores. Such approaches are suitable only for very high-resolution ($\sim$ 2000 dpi) fingerprint images and their performance is likely to be adversely affected by fingerprint degradation caused by skin conditions. 
To circumvent these challenges, Jain \textit{et al.} \cite{jain2007pores} presented a hierarchical fingerprint recognition approach that utilizes features from all the three levels. In their approach, pores are detected by applying Mexican-hat wavelet transform on the linear combination of the original and the enhanced fingerprint image. %During the matching process, firstly, level 1 features are considered and if the matching score is not greater than some threshold then the subsequent  features (level 2 and level 3) are considered. 
The fingerprints are first matched using minutiae and level-3 features are extracted in the neighbourhood of the matched minutiae points. The extracted level-3 features  are then matched using the iterative closest point (ICP) algorithm.
Later on, Zhao \textit{et al.} \cite{zhao2009direct} proposed an approach, in which the pores are extracted using the adaptive pore filtering  \cite{zhao_ICPR}. For each pore, a descriptor is formed by considering the pixel intensities in the neighbourhood. The initial correspondences are established through dot product and are refined using the random sample consensus (RANSAC) algorithm. The authors have demonstrated the usefulness of pores for biometric recognition using partial fingerprint images, which may not contain sufficient level-2 features \cite{Zhao20102833,ZHAO_partial}.  
Liu \textit{et al.} \cite{sparse_fing} proposed an improved direct pore matching approach, which employs the same pore descriptor as in \cite{zhao2009direct}. The coarse pore correspondences obtained through sparse representation are refined using the weighted RANSAC (WRANSAC) \cite{WRANSAC}. This work has been extended in \cite{LIU_PR}, which employs the tangent distance and sparse representation to compare the pores extracted from the reference and probe fingerprint images. 

Recently, Lemes \textit{et al.} \cite{Lemes} proposed a pore detection approach with a low computational cost. Their approach is adaptive and handles variations in the pore size. Firstly, a binary fingerprint image is obtained through global thresholding. For every white pixel, the average valley width is then estimated by computing the distance to neighbouring dark pixels in each of the four directions. The average valley width is used to define the size of a mask centered on each white pixel. Bright pixels inside the mask are then used to define a local threshold $T_{local}$ and a local radius $r_{local}$. Finally, a circle centered at each bright pixel with its local radius $r_{local}$ is used to determine whether the bright pixel is part of a pore or not. 
Segundo and Lemes \cite{segundo} improved the dynamic pore filtering approach \cite{Lemes} by considering the average ridge width in place of the average valley width to obtain the global and local radii, which are used in the same manner as in \cite{Lemes} to estimate the pore coordinates. 
 The authors in \cite{segundo} performed ridge reconstruction from the detected pores by employing Kruskal's minimum spanning tree algorithm. In the matching stage, a scale invariant feature transform (SIFT) based descriptor is obtained for each pore and the pores with bidirectional correspondences are used to compute the matching score. The ridge structure and ridge consistency of the corresponding pores are also used to generate the matching score.
Most recently, Dahia and Segundo \cite{CNN_SIFT} presented an approach to generate pore annotation by aligning fingerprint images in the training set, followed by   learning a descriptor for each of the pore patches by using an existing CNN-based patch matching model, HardNet \cite{HardNet2017}.
The two-step method \cite{Pore_EA} to compare fingerprint pores aligns the fingerprint images using a data-driven descending algorithm. The alignment process utilizes the fingerprint ridges and orientation field. Once the fingerprint images are aligned, the pores present in the overlapping area between the two fingerprint images are matched using a graph comparison method.

A review of the literature  indicates that there is room for improvement in level-3 feature detection and the subsequent matching. The objectives of this work is to explore deep pore-descriptors and to advance the state-of-the-art in high-resolution fingerprint recognition. To this end, we have utilized CNN-based deep learning, which has proven successful for various computer vision problems \cite{facenet,Deepface,deep_conv}. %However, the application of CNN in biometrics is still very limited especially for fingerprint recognition. The main reason could be the unavailability of the large labeled dataset required for training the CNN.
%Therefore, in this paper we have focused on the aforementioned directions.  
The key contribution of this paper is a residual  learning-based convolutional  neural network,  referred  to  as  PoreNet, that  learns distinctive feature representation from pore patches  in  high-resolution fingerprint images. In addition, we have developed an automated approach to generate labels for the pores that are common to different impressions of a finger belonging to the training set. We have also  studied the effect of cross-sensor data on the proposed approach. Specifically, this is the first study that examines the performance of a learning based fingerprint recognition approach by testing the model on cross-sensor fingerprint images. The in-house high-resolution fingerprint dataset used in this study will be made publicly available to further research in this area.
%Pore label generation is quintessential for employing CNN-based approach.
 
%Finally, we have collected an in-house high-resolution fingerprint dataset, referred to as IITI-HRF by collecting fingerprint samples from 100 subjects. And reported  results of the proposed and the existing approaches on two different datasets.

The rest of this paper is organized as follows: Section \ref{PM} presents an introduction of the proposed methodology followed by a detailed description of the pore label generation approach and the pore descriptor learning approach. Experimental results and discussion are presented in Section \ref{results}. Finally, our concluding remarks are presented in Section \ref{conclude}. 

\section{Proposed Method}
\label{PM}
\begin{figure*}[!ht]

\captionsetup[subfloat]{farskip=10pt,captionskip=1pt}
\centering
%\vspace*{-10mm}
\subfloat[Detection of pores in high-resolution fingerprint image using DeepResPore\label{fig:pore_det}]{\includegraphics[width=0.8\textwidth]{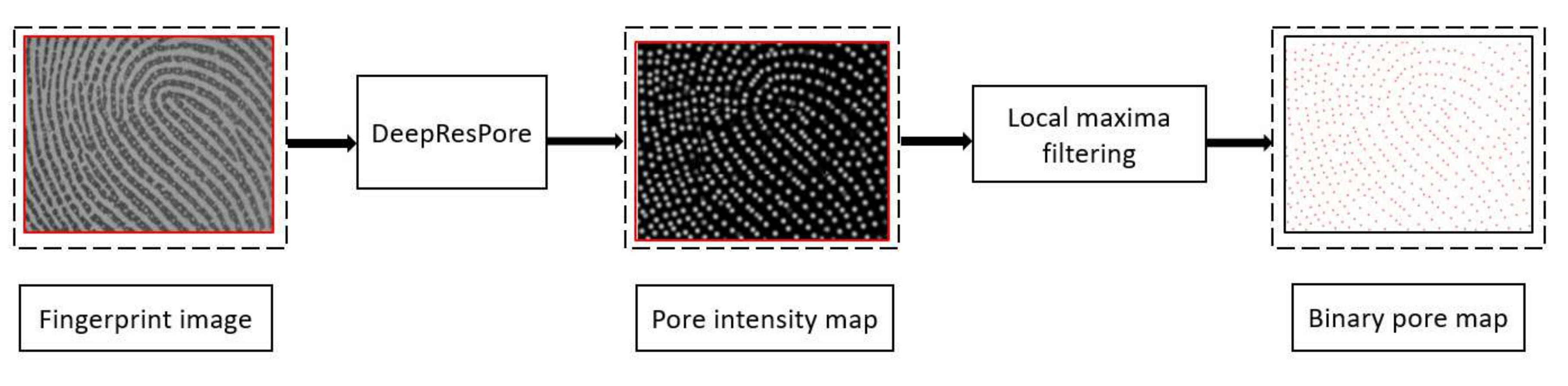}}\\ 

\subfloat[Deep pore descriptor-based fingerprint matching\label{fig:pore_mat}]{\includegraphics[width=0.8\textwidth]{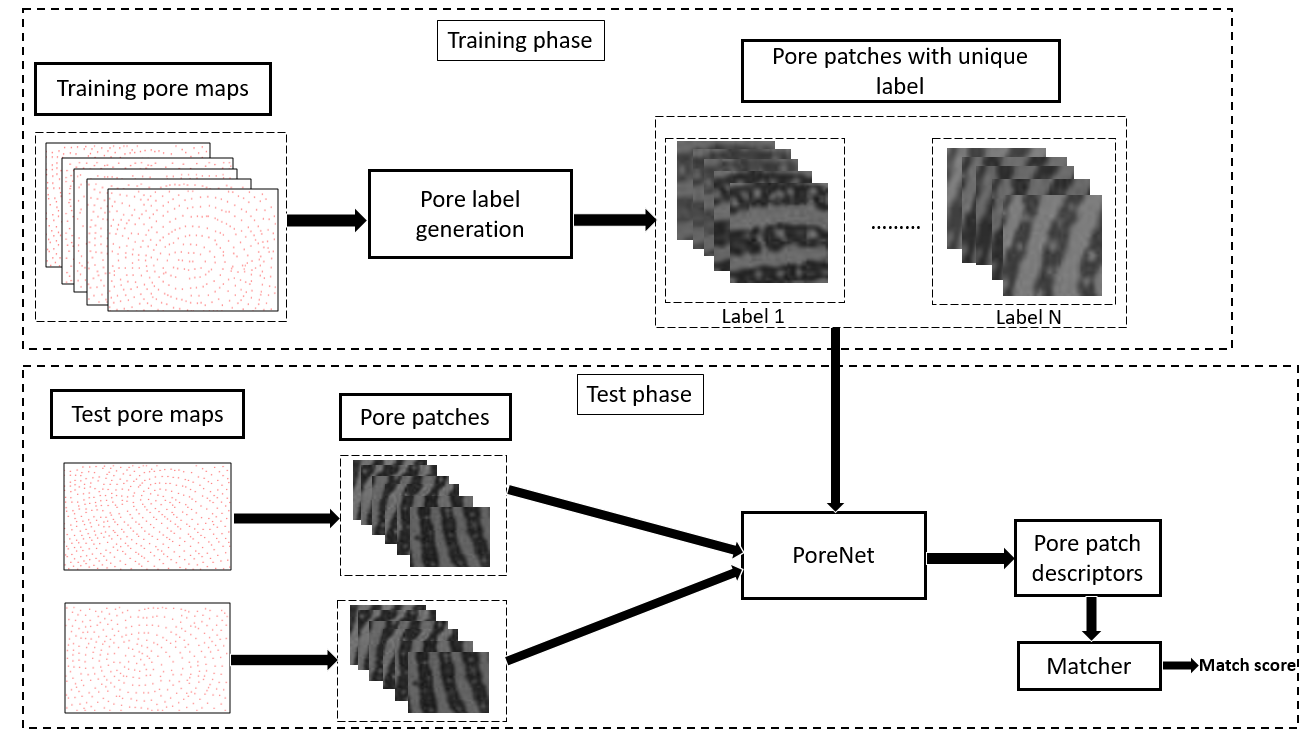}}
\caption{Schematic diagram of the proposed fingerprint recognition approach}
\label{blockdiagram}
 %\vspace*{-5mm}
\end{figure*}

The proposed method employs a CNN model,  DeepResPore \cite{vij_cnn}, to detect  pores  in  the  input  fingerprint image. It generates a  pore intensity map, which is processed to obtain a binary pore map. Thereafter, another CNN model is used to compute a deep descriptor for a patch around each detected pore. This residual  learning-based \cite{Resnet}  CNN model, referred  to as  PoreNet, has been trained to compute distinctive feature  representation  from pore  patches. For verification,  a Euclidean distance-based matching  score is generated by  comparing the pore descriptor sets obtained from the probe and reference fingerprint  images. 
%In the first stage, DeepResPore\cite{vij_cnn} is used to detect pores in the input high-resolution fingerprint image. 
%In second stage, pore descriptor is learned by employing a customized  deep residual learning-based CNN model referred to as PoreNet.
%PoreNet takes pore patches as input and converts them to the corresponding deep embeddings. An online triplet mining approach is then employed on the embeddings to select the  triplets consisting of anchor, positive and negative samples. PoreNet then learns the deep embedding representation such that it minimizes the triplet loss. 
%Finally, in the verification step, descriptors generated from the PoreNet are compared using the Euclidean distance. Further, a distance ratio $d_{r}$  is employed to refine the matched descriptor pairs. In the proposed approach, the match score is equal to the number of the matched  descriptors between the two fingerprint images. 
The schematic diagram of the proposed method is presented in Fig. \ref{blockdiagram}.

% \begin{figure*}[ht]
% \centering
% \includegraphics [height=4in,width=6.5in]{poremap_matching.eps}
% % \vspace*{-30mm}
% \caption{Schematic diagram of the proposed fingerprint recognition approach}
% \label{blockdiagram_b}
% \end{figure*}

\subsection{Pore label generation}\label{pore_label}

Given a training dataset, we first detect pores in the fingerprint images using DeepResPore model, as shown in Fig. \ref{fig:pore_det}. For a given input fingerprint image, DeepResPore generates a pore intensity map, from which a binary pore map is obtained through local maxima filtering.  
To train a CNN model in a supervised manner, one requires a labeled dataset. Specifically, to train PoreNet, we require pore patches and their corresponding labels.  To this end, we have first obtained labels for the pores that are common to different impressions of a  finger.

We have generated pore labels using a handcrafted key-point descriptor namely, DAISY \cite{DAISY}, which has been shown to be very effective in representing pores \cite{Anand2019}. % Therefore, in this work,  we have employed DAISY-based descriptor for generating the pore labels.
To obtain  DAISY descriptor for a given image $I$, 
firstly, a set of orientation maps $O_{n}$, $1\leq n \leq N$, are computed as follows \cite{DAISY}:
\begin{equation}\label{eq1}
\centering
  O_{n}(i,j)=max\Bigg(0, \frac{\partial I}{\partial n}\Bigg)
  \vspace{-0.5em}
\end{equation}
where $\frac{\partial I}{\partial n}$ is the image derivative computed at $(i,j)$ along the direction
$n$. 
The orientation maps $O_n$ are then convolved with a set of Gaussian kernels having different standard deviation ($\sigma$) values. Each pixel's neighborhood in the convolved orientation map is then divided into overlapping circular regions on a series of concentric rings around the center pixel.  Next, a normalized histogram of values from the convolved orientation maps is computed for each of the circular regions.  Finally, the histograms computed from each of the circular regions are concatenated to obtain  the DAISY descriptor.  A detailed description of the DAISY descriptor can be found in \cite{DAISY}.

The steps involved in our pore-label generation process are as follows: firstly, we identify a reference fingerprint image for each of the fingers present in the training dataset. The objective is to select a reference fingerprint image that has the maximum number of pores common to all other impressions of that finger.
%For a given finger, we have several impressions in the training dataset. We need to obtain the reference fingerprint image for each of the finger present in the training dataset
To this end, we treat the detected pore coordinates as key-points and compute DAISY-based descriptor for each of the detected pores. Since the training set contains multiple impressions of a finger, we consider one impression ($i$) at a time and   perform pair-wise comparisons with rest of the impressions ($j$) using the DAISY-based pore descriptor. The  match score (i.e. the number of matched descriptors) generated from each comparison is stored in $S_{ij}$. %We then sum all the $S_{ij}$ to obtain final match score corresponding to an impression $i$.
This is repeated for every impression of a finger and the index $R$ of the reference fingerprint image is determined as follows:
\begin{equation}
\label{err1}
 R=\operatorname*{arg\,max}_{i \in\{1, 2, \hdots, N\}} \bigg (\sum_{j=1, j \neq i}^{N}S_{ij} \bigg ) 
\end{equation}
where $N$ is the number of impressions of a finger. 
%and $N=(r-1)$ is total number of comparisons made.
The impression, for which the sum of match scores is maximum, is considered to be the reference image.

Once the reference fingerprint image is identified, the remaining impressions are aligned with the reference fingerprint to find the common pores that are present in all the impressions of that finger. The affine transformation \cite{Hartley:2003:MVG:861369, TORR_MLESAC} for alignment is estimated using pore correspondences established based on the earlier comparison of DAISY-based pore descriptors.
%Firstly, we obtain the DAISY-based pore descriptors of the reference and the moving fingerprint images.
%Further, the Diasy-based descriptors of the reference and the moving image are matched to obtain the common pore points. 
%and a transformation matrix $T_M$ is  estimated based on the matched descriptors \cite{Hartley:2003:MVG:861369, TORR_MLESAC}. Furthermore, $T_M$ is applied on the moving fingerprint image to obtain the transformed image which has overlapping region with the reference image.
After aligning all impressions of a finger with its reference image, we identify the reference fingerprint image pore coordinates for which there are corresponding pores within the image boundaries of every transformed impression. We consider them to be common pores. 
%we consider each pore coordinates in the reference fingerprint and find weather it is present in all other transformed impressions by checking the pixel value of the corresponding coordinates in the transformed image.
At the end of this process, we have the coordinates of the common pores that are present in all impressions of each finger. Finally, a patch of $41\times 41$ pixels centered at each of the common pores is extracted and assigned a unique label. It is important to note that a finger having $P$ pores common to all its $r$ impressions will have $P\times r$ pore patches having $P$ unique labels in the training set.
%A detailed description of the pore label generation approach is presented in  Algorithm \ref{algo1}.

% \begin{algorithm}
% %\SetAlgoLined
% \caption{\textcolor{blue}{Pore label generation}}
% \label{algo1}
% \begin{algorithmic}[1]
%  \renewcommand{\algorithmicrequire}{\textbf{Input:}}
%  \renewcommand{\algorithmicensure}{\textbf{Output:}}
% \REQUIRE{Training fingerprint images}
% \ENSURE{Pore patches $P_{patch}$ with corresponding labels $P_{label}$ }
% \STATE Apply DeepResPore on the given fingerprint image to detect pores $P$ present in it\\
%  \FOR {each pore $p \leftarrow 1 $ \TO $P$}
% \STATE {Obtain DAISY-based pore descriptors, [${PD_{1},\ldots, PD_{Tp}}$] for all the extracted pores from the fingerprint image, \\where
%  $Tp=$ total number of pores extracted}
%   \ENDFOR
%  \STATE Repeat steps 1 to 4 for all impressions of a finger
% \STATE Obtain the reference fingerprint index $R$ for each finger using eqn. \ref{err1} 
% \STATE Align all other fingerprint impressions of same finger with the reference image
% \STATE Find common pores between the aligned images
% \STATE Create a patch of size $41\times 41$ pixels centered at each common pore
% \STATE  Assign unique label to pore patches of each common pore 
% \STATE  Repeat steps 6 to 10 for all the fingers present in training dataset
% \STATE Return pore patches $P_{patch}$ with corresponding labels $P_{label}$
%   \end{algorithmic}
%   \end{algorithm}
\begin{figure*}[ht]
\centering
\includegraphics[width=0.9\textwidth]{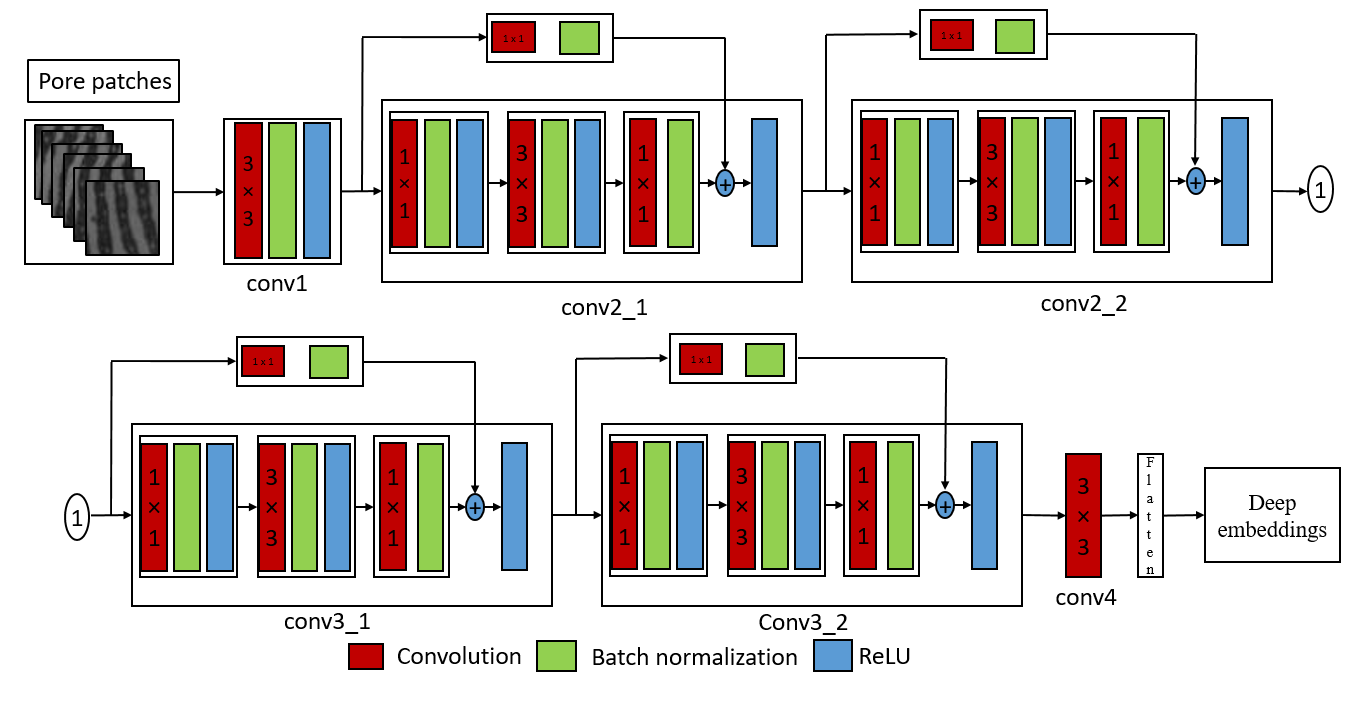}
% \vspace*{-30mm}
\caption{Schematic diagram of PoreNet showing generation of deep embedding from pore patches} 
\label{blockdiagram_porenet}
\end{figure*}

\subsection{Pore descriptor learning}
{On completion of the pore-label generation process,  we have a set of pore patches $X=\{x_{1},x_{2},\ldots,x_{T}\}$ and the corresponding set of labels $Y=\{y_{1},y_{2},\ldots,y_{T}\}$ for each of the fingerprint images in the training set. To learn a pore-patch descriptor,  we have designed  a  customized residual learning-based CNN model, referred to as PoreNet.}
%\textcolor{blue}{We have developed a residual learning-based CNN model for the following reasons. Firstly, the residual learning provides a large receptive field \cite{Luo:2016}, which is expected to result in increased representational power suitable for pixel-level predictions. Secondly, the residual learning leads to faster convergence and yields less training error compared to the plain stacked convolutional layers \cite{Resnet}.} 
 { The detailed architecture of PoreNet is presented in Table \ref{CNN}. As can be seen, the proposed model has 14 learnable layers with 4 residual blocks consisting of 4 shortcut connections. %Specifically, we have employed $1\times1$ convolutional and the identity shortcut connections in an alternating manner.
 The PoreNet takes a pore-patch $I_{p}$ of size $41\times41$ pixels as input and converts it into a 1681-dimensional pore descriptor. As shown in Fig. \ref{blockdiagram_porenet}, the PoreNet does not perform any pooling operation in order that the size of the output feature map remains the same as that of the input. %Further, pooling operation could have adversely affected the pore patch representation.
 }
At the end of the network, a convolutional layer  containing a single filter is introduced to generate the final feature map, which is flattened and normalized such that $l_2$-norm of the output embedding is equal to one \cite{facenet}.
All convolution operations in   PoreNet are performed with a stride of one. At every stage, zero padding is employed to maintain the size of the feature map. 
Furthermore, all convolutional layers, except the last one, perform convolution  followed by batch normalization \cite{BN}  and ReLU activation.

In the training phase, the PoreNet is trained from the scratch. It is trained end-to-end in a supervised learning manner with the objective of minimizing the value of a triplet loss function. A batch of pore patches are fed to PoreNet, which generates the corresponding high-dimensional embeddings. Next, an online triplet mining scheme namely, batch-hard triplet mining \cite{facenet}, is applied to the obtained  embeddings. 
In this method, for each anchor sample $(a)$, we first obtain the hardest positive sample $(p)$  having the same label as that of the anchor and the hardest negative sample $(n)$ having a label different from the anchor. Finally, we form the following  triplet loss \cite{facenet}:
\begin{equation}\label{loss}
  L_{triplet}= \max \big \{ d(a,p) - d(a,n) + margin , 0\big \}
\end{equation}
where $d(a,p)$ is the distance between $a$ and $p$ and $d(a,n)$ is the distance between $a$ and $n$.
The PoreNet is trained in an end-to-end manner to minimize the above triplet loss.

\begin{table}[!ht]
\centering
\caption{Detailed architecture of PoreNet}
\label{CNN}
\begin{tabular}{|c|c|c|}
\hline
\textbf{Layer name} & \textbf{Output shape} & \textbf{Kernel}                                                                        \\ \hline

conv1      & $41\times41, 16$       & $3\times3$, $16$, stride 1, padding same                                                            \\ \hline
conv2\_x   &$41\times41, 64$      & \Bigg[\begin{tabular}[c]{@{}c@{}}$1\times1$, $32$ \\ $3\times3$, $32$ \\ $1\times1$, $64$\end{tabular}\Bigg] $\times 2$                  \\ \hline
conv3\_x   & $41\times41, 128$      & \Bigg[\begin{tabular}[c]{@{}c@{}}$1\times1$, $64$ \\ $3\times3$, $64$\\ $1\times1$, $128$ \end{tabular}\Bigg] $\times 2$                    \\ \hline
% conv4\_x   &$80\times80$      & \Big[\begin{tabular}[c]{@{}c@{}}$3\times3$, $256$ \\ $3\times3$, $256$\end{tabular}\Big]  $\times2$                            \\ \hline
% conv5\_x   &$80\times80$      & \Big[\begin{tabular}[c]{@{}c@{}}$3\times3$, $512$\\  $3\times3$, $512$\end{tabular}\Big]   $\times2$                           \\ \hline
conv4      & $41\times41, 1$       & $3\times3$, $1$                                                                            \\ \hline
Flatten      & $1681$       & $-$                                                                            \\ \hline
$l_2$-norm      & $1681$       & $-$                                                                            \\ \hline
Total parameters & 142,881 & $-$                                                                                    \\ \hline
\end{tabular}
%\vspace*{5mm}
\end{table}

Once the PoreNet is trained, the task  is to match a pair of fingerprint images using the generated pore-patch descriptors.
Considering two fingerprint images $I_1$ and $I_{2}$ containing $n$ and $m$ number of pore patches, respectively, the PoreNet will generate two sets of pore descriptors $P_1 \in \mathds{R}^{n\times 1681}$  and $P_2 \in \mathds{R}^{m\times 1681}$. The pore descriptors in $P_1$ are compared with those in $P_2$ using the Euclidean distance and the pairs of descriptors matched bidirectionally are retained. Finally, the matches are  refined using a distance ratio ($d_{r}$) criterion involving the distance to the second nearest neighbour \cite{Lowe_sift}.    

\begin{table*} [!ht]
\caption{Details of the PolyU HRF dataset}
\label{DB_table}
\centering
\begin{tabular}{|>{\raggedright\arraybackslash}M{2cm}|M{2cm}|M{2cm}|c|M{3cm}|c|}
\hline
\multicolumn{1}{|>{\centering\arraybackslash}M{2cm}|}{Dataset}&Resolution (dpi)&Image size (pixels)&Fingers&Images per finger per session&Total images\\
\hline
 DBI:Training set&1200 &$320\times 240$&35& 3 &210\\
\hline
 DBI:Test set&1200 &$320\times 240$&148& 5 &1480\\
\hline
 DBII&1200 &$640\times 480$&148& 5 &1480\\
\hline
\end{tabular}
%\end{center}
\end{table*}

\section{Experimental results and discussion}
In this section, we first present the details of preparation of the dataset for training the PoreNet model,  followed by results of our experiments. 
\label{results}

\subsection{Dataset preparation}
We have performed experiments on the publicly available PolyU HRF dataset \cite{POLYU}, which contains high-resolution fingerprint images of resolution 1200 dpi in two different sets DBI and DBII. DBI contains two subsets: DBI-train containing 210  partial fingerprint images from 35 fingers, each contributing 6 impressions and DBI-test containing 1480 partial fingerprint images. The size of these partial fingerprint images is $320\times 240$ pixels. On the other hand, DBII contains 1480 complete fingerprint images of size $640\times480$ pixels. Fingerprints images in DBI-test and DBII are collected from 148 fingers in two different sessions, with each finger contributing 5 impressions per session. The detailed description of PolyU HRF dataset is provided in Table \ref{DB_table}. While the pore patches for training PoreNet have been obtained from DBI-train, the datasets DBI-test and DBII have been used to evaluate the proposed approach using the experimental protocol adopted in the previous works  \cite{segundo}, \cite{CNN_SIFT}, \cite{Pore_EA}. A training set $(80\%)$ and a validation  set $(20\%)$ have been obtained by randomly partitioning DBI-train. We have obtained pore patches from every image in the training set and generated their  labels using the method detailed in Section \ref{pore_label}.  We have also employed data augmentation techniques to increase the amount of training data. Specifically, we have generated twenty (through ten rotations with angles randomly selected between $-20^{\circ}$ and $20^{\circ}$ and ten translations with shifts randomly selected between $-5$ and $5$ pixels) geometrically transformed images from every training sample. Further, we have generated an additional ten images by varying the contrast using the gamma transformation with gamma value selected in the range 0.45 to 0.9 in steps of 0.05. Overall, we have generated 815,982 labelled pore patches to train the PoreNet.

\subsection{Experimental results}
\label{ER}
We have trained  PoreNet for 100 epochs using a batch size of 256 and learning rate of 0.0001. The loss function has been optimized using the adaptive moment estimation (ADAM) \cite{ADAM}. The margin for the triplet loss function was set to 0.8. All our experiments have been performed on a computer with 3.60 GHz Intel core i7-6850K processor, 48 GB RAM and Nvidia GTX 1080 8 GB GPU. While the pore patches have been generated in MATLAB environment, the PoreNet has been trained and tested using TensorFlow \cite{Tensorflow} in Python environment.
% backend with Keras \cite{chollet2015keras} library. 

\begin{table}[ht!]
\caption{EERs on PolyU datasets }
\centering
\begin{tabular}{|M{3cm}|M{2cm}|M{2cm}|} \hline
\multirow{2}{*}{Method}&\multicolumn{2}{|c|}{EER (\%)}\\ \cline{2-3}
&DBI& DBII \\ \hline
 Jain \textit{et al.} \cite{jain2007pores}&$30.45\%$&$7.83\%$\\ \hline
 Zhao \textit{et al.} \cite{zhao2009direct}&$15.42\%$&$7.05\%$\\ \hline
 Liu \textit{et al.} \cite{sparse_fing} & $6.59\%$&$0.97\%$ \\ \hline
 Liu \textit{et al.} \cite{LIU_PR} & $3.25\%$&$0.53\%$ \\ \hline
Segundo and Lemes \cite{segundo}& $3.74\%$ & $0.76\%$ \\ \hline
Dahia and Segundo \cite{CNN_SIFT}& $4.18\%$ & {1.14\%} \\
\hline
Xu \textit{et al.} \cite{Pore_EA}& \textbf{1.73\%} & \textbf{0.51\%} \\ \hline
Proposed approach & $2.91\%$ & $0.57\%$ \\ \hline
\end{tabular}
\label{results_polyu}
\vspace*{3mm}
\end{table}

\begin{figure}[!h]
\captionsetup[subfloat]{farskip=10pt,captionskip=1pt}
\centering
%\vspace*{-10mm}
\subfloat[DBI]{\includegraphics[height = 5cm, width=0.5\textwidth]{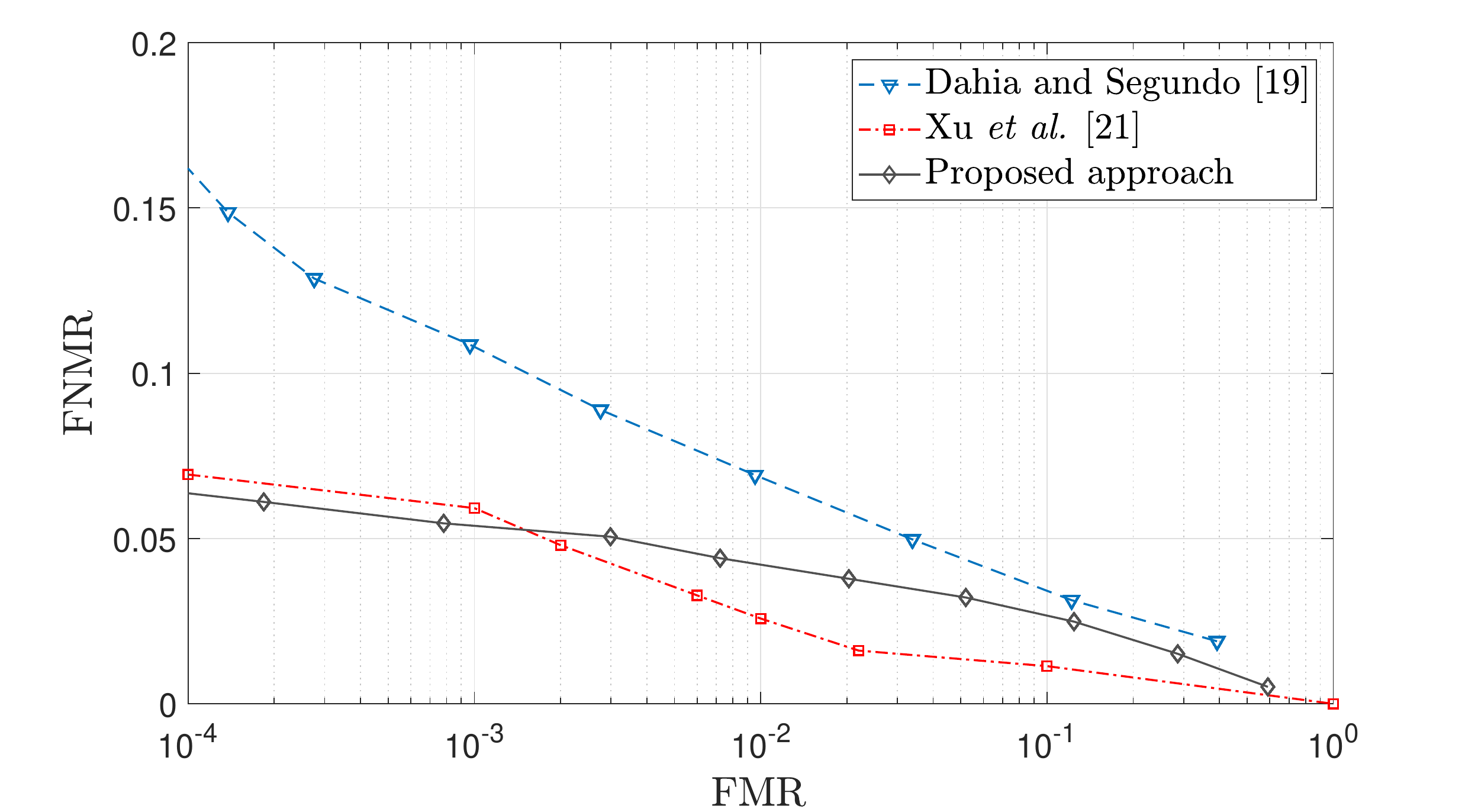}}\\

\subfloat[DBII]{\includegraphics[height = 5cm, width=0.5\textwidth]{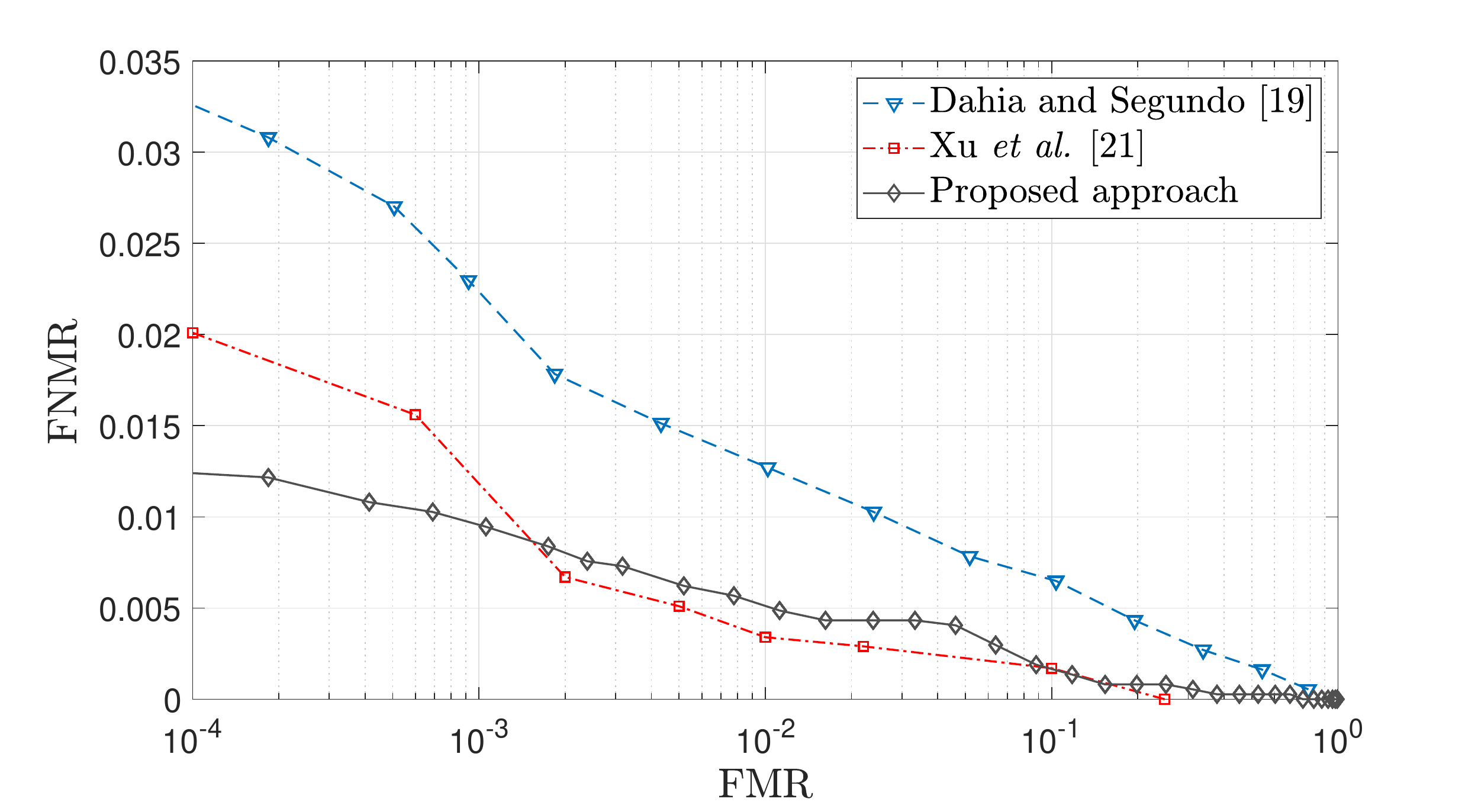}}
\caption{A comparative analysis using DET curves}
\label{ROC_DBI}
\end{figure}

To make a fair comparison with the existing approaches, we have followed the same experimental protocol as in \cite{segundo,CNN_SIFT,Pore_EA}. A set of genuine scores has been obtained by comparing every fingerprint image from the second session with all fingerprint images belonging to the same finger collected in the first session. On the other hand, impostor scores have been generated by comparing the first sample of every finger from the second session with the first sample of all other fingers from the first session. Thus, we have a total of 3700 $(148\times 25)$ genuine scores and 21,756 $(148\times 147)$ impostor scores. We report the following performance metrics: equal error rate (EER), FMR1000 and FMR10000 \cite{FMR_EER}. In addition, we
present the detection error trade-off (DET) curve to help ascertain the performance of the proposed approach. The $d_r$ value was empirically set to 0.8 using the validation set. Table \ref{results_polyu} \footnote{The EERs of the existing approaches \cite{jain2007pores}, \cite{zhao2009direct},\cite{sparse_fing}, \cite{segundo} are taken directly from the results presented in \cite{segundo} and those of the approaches \cite{LIU_PR} and \cite{Pore_EA} are taken from \cite{Pore_EA}. EERs values of \cite{CNN_SIFT} are obtained from the source code provided by the authors.} presents EERs of the proposed and the existing approaches on PolyU datasets.
Since the recent methods show comparable performance in terms of the EER, we have further analysed their performance using FMR10000 and FMR1000. These results are presented in Table \ref{FMR_values}. As can be seen, the proposed approach achieves lower FNMRs on both the datasets. To ascertain this superior performance, we have plotted DET curves (please see Fig. \ref{ROC_DBI}). 
These curves clearly show that the proposed approach achieves better FNMR for low FMRs, specifically, FMR in the range $10^{-4}$ to $10^{-3}$. The performance improvement in this region translates into increased security and convenience for users.
  %Specifically, the proposed PoreNet model provides an improvement of  $0.47\%$, $0.5\%$ and $0.77\%$, $0.35\%$ points in FMR10000 and FMR1000 values  over the current state-of-the-art approach  on DBI and DBII, respectively.
Furthermore, the proposed approach requires on average $1.41$ seconds and $2.80$ seconds to compare a pair of fingerprint images from DBI and DBII, respectively. On the other hand, the existing approach \cite{Pore_EA}  takes $8.46$ seconds to compare a pair of fingerprint images.

Overall, the experimental results presented in this section indicate that the proposed PoreNet model provides improvement in performance over the current state-of-the-art approaches. Specifically, it achieves lower FMR10000 and FMR1000 on both benchmark datasets.

\begin{table}[ht!]
\caption{A comparative analysis using FMRs}
\centering
\begin{tabular}{|M{2.5cm}|M{1cm}|M{1cm}|M{1cm}|M{1cm}|} \hline
\multirow{2}{*}{Method}&\multicolumn{2}{|c|}{FMR10000 }&\multicolumn{2}{|c|}{FMR1000 }\\ \cline{2-5}
&DBI& DBII & DBI & DBII \\ \hline
 Dahia and Segundo \cite{CNN_SIFT}&$16.20\%$&$3.33\%$ & $10.82\%$ & $2.25\%$ \\ \hline
 Xu \textit{et al.} \cite{Pore_EA}&$6.94\%$&$2.01\%$ & $5.92\%$ & $1.31\%$\\ \hline
 Proposed approach &\textbf{6.47\%}& \textbf{1.24\%} & \textbf{5.42\%} & \textbf{0.96\%}\\ \hline
\end{tabular}
\label{FMR_values}
 \vspace*{5mm}
\end{table}

\subsection{Performance in a cross-sensor scenario}
\label{DA}
In this section, we present results of the experiments conducted to study the effect of cross-sensor fingerprint data on the performance of the proposed approach. As discussed previously, our model has been trained on PolyU HRF dataset \cite{POLYU}. The in-house IITI-HRF high-resolution fingerprint (1000 dpi resolution) dataset \cite{Anand2019}, which has been expanded to include
fingerprint images of 100 subjects, is used for testing. The expanded IITI-HRF high-resolution fingerprint dataset contains images of 8 fingers (all fingers except the little fingers on both the hands), each contributing 8 impressions. These images were acquired using the commercially available Biometrika HiScan-Pro fingerprint scanner. IITI-HRF dataset is partitioned into two subsets. The first one contains partial fingerprint images of size $320\times240$ pixels, while the second one contains full fingerprint images of size $1000\times1000$ pixels. The details of IITI-HRF dataset is presented in Table \ref{IITI_table}. 

\begin{table} [!ht]
\caption{Details of the IITI-HRF dataset}
\label{IITI_table}
\centering
\begin{tabular}{|>{\raggedright\arraybackslash}M{1.5cm}|M{1cm}|M{1.5cm}|c|c|}
\hline
\multicolumn{1}{|>{\centering\arraybackslash}M{1.5cm}|}{Dataset}&Resolution (dpi)&Image size (pixels)&Fingers&Total images\\
\hline
 IITI-HRFP&1000 &$320\times 240$&800&6400\\
\hline
 IITI-HRFC&1000 &$1000\times 1000$&800&6400\\
\hline
\end{tabular}
%\end{center}
\end{table}

\begin{table}[h]
\caption{EERs on IITI-HRF dataset }
\centering
\begin{tabular}{|M{3cm}|M{2cm}|M{2cm}|} \hline
\multirow{2}{*}{Method}&\multicolumn{2}{|c|}{EER (\%)}\\ \cline{2-3}
&IITI-HRFP& IITI-HRFC \\ \hline
 
Segundo and Lemes \cite{segundo}& $8.37\%$ & $2.90\%$ \\ \hline
Proposed approach & $9.58\%$ & $4.28\%$ \\ \hline
\end{tabular}
\label{results_IITI}
\end{table}

\begin{figure}[!h]
\captionsetup[subfloat]{farskip=10pt,captionskip=1pt}
\centering
%\vspace*{-10mm}
\subfloat[IITI-HRFP]{\includegraphics[ width=0.5\textwidth]{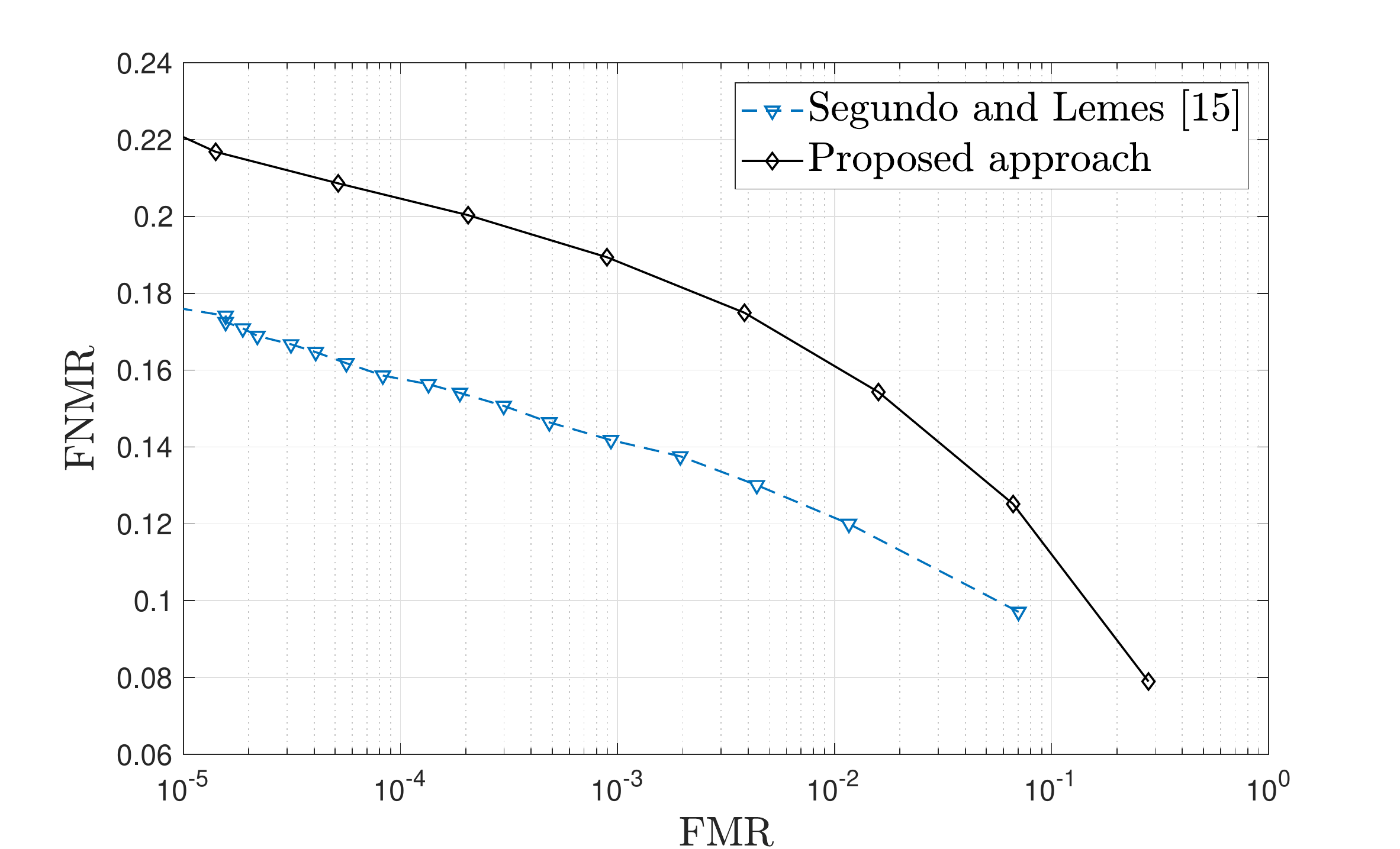}}\\

\subfloat[IITI-HRFC]{\includegraphics[ width=0.5\textwidth]{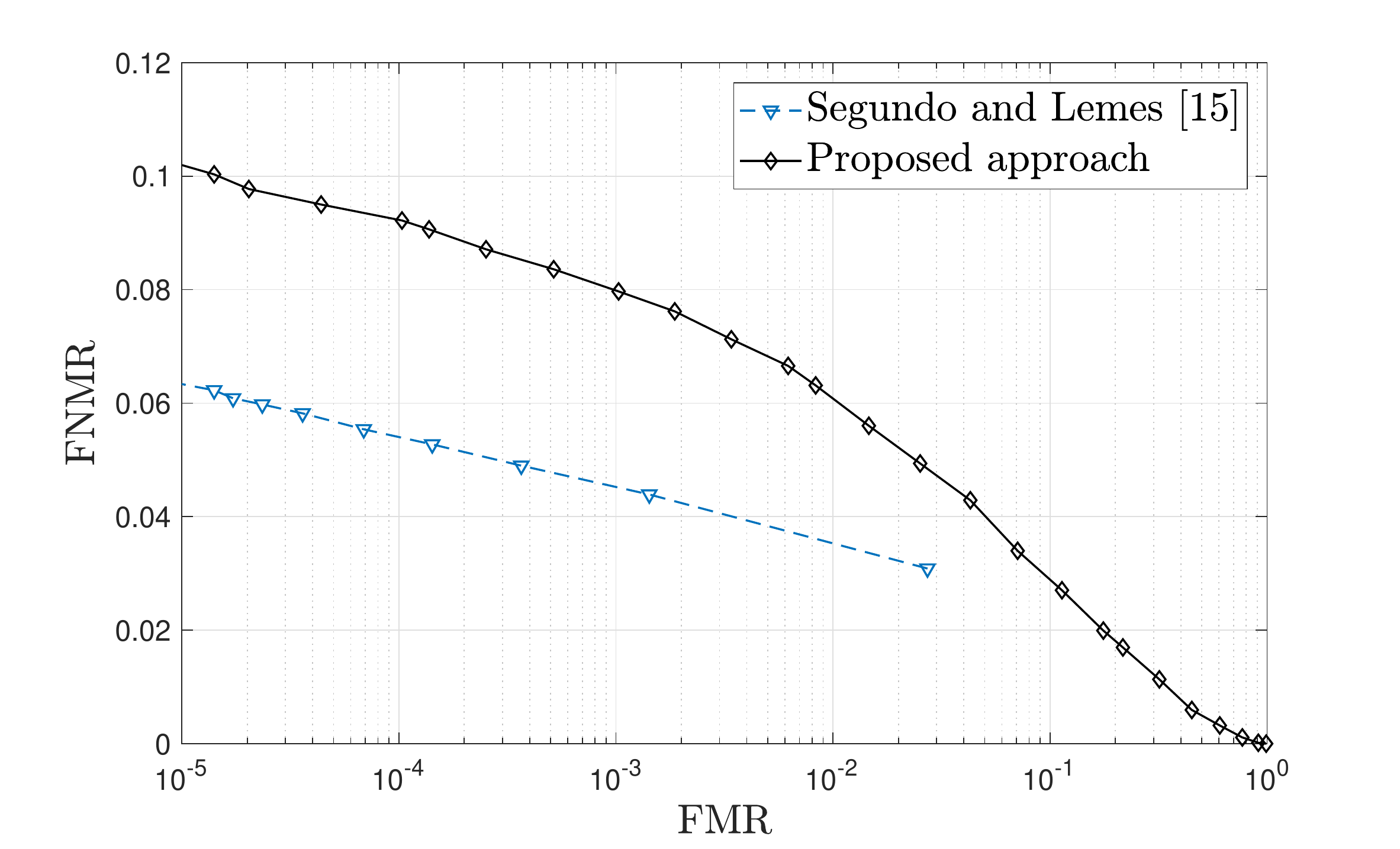}}
\caption{Cross-sensor performance comparison using DET curves}
\label{ROC_IITI}
\end{figure}
To make a comparison, we have also evaluated the performance of the hand-crafted feature based approach  proposed by Segundo and Lemes \cite{segundo} on IITI-HRF dataset.
 For both these approaches, genuine and the impostor scores on IITI-HRF dataset have been generated using the following protocol: for each finger, 16 genuine scores have been generated by comparing the last four impressions with each of the first four impressions of the same finger. On the other hand, impostor scores have been generated by comparing the fifth impression of each finger with the first impression of all other fingers. This way, we have a total of 12,800 ($800\times16$) genuine scores and 639,200 ($800\times799$) impostor scores. The EERs and DET curves of both the approaches are presented in Table \ref{results_IITI} and  Fig. \ref{ROC_IITI}, respectively.  As can be observed, the hand-crafted feature-based approach \cite{segundo} provides lower EER values as compared to the proposed learning-based approach  on IITI-HRFP as well as IITI-HRFC.  This may be due to the fact that our approach has been trained on PolyU HRF dataset and there exists a considerable variation in terms of resolution and the quality of pores between fingerprint images belonging to PolyU HRF  and IIT-HRF. Therefore, the proposed learning based approach appears to suffer from the problem of domain adaptability.

The results presented in this section underline the key challenge facing learning-based fingerprint recognition approaches, specifically, to overcome domain variability in cross-sensor scenarios. Employing domain adaptation techniques such as the one proposed in \cite{Deep_DA} in the training phase is likely to improve the cross-sensor performance of the learning-based approaches.

\section{Conclusion}
\label{conclude}
In this paper, we have presented a deep pore-descriptor based method for high-resolution fingerprint recognition. Specifically, we have
developed a residual learning-based CNN named PoreNet to learn a descriptor from pore patches in fingerprint images. The trained PoreNet generates deep embeddings from a given fingerprint image. To train PoreNet efficiently, we have also developed a method that  generates pore labels by transforming the images based on the matched DAISY-based pore descriptors and finding the common pores. 
The results of our evaluations on the benchmark PolyU HRF datasets demonstrate the effectiveness of the proposed PoreNet in generating pore descriptors for high-resolution fingerprint recognition.
Most importantly, the proposed PoreNet model achieves state-of-the-art performance in terms of FMR10000 and FMR1000. In future, we plan to work on the domain adaptability of the PoreNet model for cross-sensor fingerprint matching. 

 \section*{Acknowledgment}
 We thank Maur\'icio P. Segundo for sharing the source codes of their approach. We also thank Yuanrong Xu for providing us with the data points of the DET curve of their approach.
%\pagebreak

\ifCLASSOPTIONcaptionsoff
  \newpage
\fi



%\bibliographystyle{IEEEtran}
%\begin{thebibliography}{1}


%\end{thebibliography}

\bibliographystyle{IEEEtran}
%\bibliography{submission_example}
%\bibliographystyle{abbrv}
\bibliography{IEEEabrv,vijay_pore_match}
%\end{document}


%\end{IEEEbiography}


%\newpage



%\vfill


\end{document}